\documentclass[letterpaper]{article}

\usepackage{aaai} 
\usepackage{times}
\usepackage{helvet}
\usepackage{courier}
\usepackage{graphicx}
\usepackage{algorithm}
\usepackage{algorithmic}
\usepackage{amssymb}
\usepackage{amsmath}
\usepackage{wasysym}
\usepackage{amsthm}
\usepackage{stmaryrd}
\usepackage{natbib}

\author{Joel Veness, Marc G. Bellemare, Marcus Hutter, Alvin Chua, Guillaume Desjardins\\
Google DeepMind, Australian National University\\
\texttt{\{veness,bellemare,alschua,gdesjardins\}@google.com}\\
\texttt{marcus.hutter@anu.edu.au}\\
}

\begin{document}

\title{Compress and Control}
\maketitle


\newif\ifextended


\newcommand{\cnc}{{\sc cnc}}
\newcommand{\pong}{{\sc Pong}}
\newcommand{\skipcts}{{\sc SkipCTS}}
\newcommand{\lempelziv}{{\sc Lempel-Ziv}}
\newcommand{\sad}{{\sc sad}}
\def\fr#1#2{{\textstyle{#1\over#2}}} 
\newcommand{\nats}{\mathbb{N}}
\newcommand{\cX}{\mathcal{X}}
\newcommand{\cdbar}{\,|\,}
\newcommand{\cbar}{\,|\,}
\newcommand{\dens}{{\it Density}}
\newcommand{\cA}{{\cal A}}
\newcommand{\cO}{{\cal O}}
\newcommand{\cR}{{\cal R}}
\newcommand{\cC}{{\cal C}}
\newcommand{\cP}{\mu}
\newcommand{\E}{\mathbb{E}}
\newcommand{\cSP}{{\cal P}}
\newcommand{\searchalg}{{$\rho$UCT}}
\newcommand{\lb}{\llbracket}
\newcommand{\rb}{\rrbracket}
\newcommand{\bstr}[1]{\llbracket #1 \rrbracket}
\newcommand{\mytick}{\ding{51}}
\newcommand{\mycross}{\ding{55}}
\newcommand{\cK}{\mathcal{K}}
\newcommand{\cM}{\mathcal{M}}
\newcommand{\cS}{\mathcal{S}}
\newcommand{\cI}{\mathcal{I}}
\newcommand{\cT}{\mathcal{T}}
\newcommand{\cD}{\mathcal{D}}
\newcommand{\cW}{\mathcal{W}}
\newcommand{\cU}{\mathcal{U}}
\newcommand{\cY}{\mathcal{Y}}
\newcommand{\cB}{\mathcal{B}}
\newcommand{\rmin}{r_{\sc min}}
\newcommand{\rmax}{r_{\sc max}}
\newcommand{\cZ}{\mathcal{Z}}
\newcommand{\SetN}{\mathbb{N}}
\newcommand{\prob}{\mathbb{P}}
\def \rhos {\rho_{\textsc{s}}}
\def \rhoz {\rho_{\textsc{z}}}

\newtheorem{defn}{Definition}
\newtheorem{lem}{Lemma}
\newtheorem{thm}{Theorem}
\newtheorem{notation}{Notation}
\newtheorem{thmapdx}{Theorem}
\newtheorem{propapdx}{Proposition}
\newtheorem{lemapdx}{Lemma}
\newtheorem{assumption}{Assumption}
\newtheorem{corollary}{Corollary}

\begin{abstract}
This paper describes a new information-theoretic policy evaluation technique for reinforcement learning.
This technique converts any compression or density model into a corresponding estimate of value.
Under appropriate stationarity and ergodicity conditions, we show that the use of a sufficiently powerful model gives rise to a consistent value function estimator.
We also study the behavior of this technique when applied to various Atari 2600 video games, where the use of suboptimal modeling techniques is unavoidable.
We consider three fundamentally different models, all too limited to perfectly model the dynamics of the system. 
Remarkably, we find that our technique provides sufficiently accurate value estimates for effective on-policy control.
We conclude with a suggestive study highlighting the potential of our technique to scale to large problems.
\end{abstract}

\section{Introduction}

Within recent years, a number of information-theoretic approaches have emerged as practical alternatives to traditional machine learning algorithms. 
Noteworthy examples include the compression-based approaches of \citet{Frank00textcategorization} and \citet{Bratko06spamfiltering} to classification, and \citet{Cilibrasi05clusteringby} to clustering.
What differentiates these techniques from more traditional machine learning approaches is that they rely on the ability to compress the raw input, rather than  combining or learning features relevant to the task at hand.
Thus this family of techniques has proven most successful in situations where the nature of the data makes it somewhat unwieldy to specify or learn appropriate features.
This class of methods can be formally justified by appealing to various notions within algorithmic information theory, such as Kolmogorov complexity \citep{li-vitanyi}.
In this paper we show how a similarly inspired approach can be applied to reinforcement learning, or more specifically, to the tasks of policy evaluation and on-policy control.

Policy evaluation refers to the task of estimating the value function associated with a given policy, for an arbitrary given environment.
The performance of well-known reinforcement learning techniques such as policy iteration \citep{howard1960dynamic}, approximate dynamic programming \citep{bertsekas1996,Powell} and actor-critic methods \citep{Sutton:1998}, for example, all crucially depend on how well policy evaluation can be performed.
In this paper we introduce a model-based approach to policy evaluation, which transforms the task of estimating a value function to that of learning a particular kind of probabilistic state model.

To better put our work into context, it is worth making the distinction between two fundamentally different classes of model based reinforcement learning methods. 
\emph{Simulation based} techniques involve learning some kind of forward model of the environment from which future samples can be generated.
Given access to such models, planning can be performed directly using search.
Noteworthy recent examples include the work of \citet{doshi09}, \citet{walsh10}, \citet{veness10}, \citet{veness10b}, \citet{AsmuthL11}, \citet{guez12}, \citet{HamiltonFP13} and \citet{covertreerl}.
Although the aforementioned works demonstrate quite impressive performance on small domains possessing complicated dynamics, scaling these methods to large state or observation spaces has proven challenging.
The main difficulty that arises when using learnt forward models is that the modeling errors tend to compound when reasoning over long time horizons \citep{talvitie14}.

In contrast, another family of techniques, referred to in the literature as \emph{planning as inference}, attempt to side-step the issue of needing to perform accurate simulations by reducing the planning task to one of probabilistic inference within a generative model of the system.
These ideas have been recently explored in both the neuroscience \citep{botvinick12,botvinick12b} and machine learning \citep{Attias03planningby,Poupart:2011:ELO:2034396.2034517} literature.
The experimental results to date have been somewhat inconclusive, making it far from clear whether the transformed problem is any easier to solve in practice.
Our main contribution in this paper is to show how to set up a particularly tractable form of inference problem by generalizing compression-based classification to reinforcement learning.
The key novelty is to focus the modeling effort on learning the stationary distribution of a particular kind of augmented Markov chain describing the system, from which we can approximate a type of \emph{dual representation} \citep{07adprl-dualrl,wang08} of the value function.
Using this technique, we were able to produce effective controllers on a problem domain orders of magnitude larger than what has previously been addressed with simulation based methods.

\section{Background}

We start with a brief overview of the parts of reinforcement learning and information theory needed to describe our work, before reviewing compression-based classification.

\subsection{Markov Decision Processes}
\label{sec:background_mdps}

A Markov Decision Process (MDP) is a type of probabilistic model widely used within reinforcement learning \citep{Sutton:1998,szepesvari10} and control \citep{bertsekas1996}.
In this work, we limit our attention to finite horizon, time homogenous MDPs whose action and state spaces are finite.
Formally, an MDP is a triplet $(\cS, \cA, \cP)$, where $\cS$ is a finite, non-empty set of states, $\cA$ is a finite, non-empty set of actions and $\cP$ is the transition probability kernel that assigns to each state-action pair $(s,a) \in \cS \times \cA$ a probability measure $\cP(\cdot \cbar s,a)$ over $\cS \times \mathbb{R}$.
$\cS$ and $\cA$ are known as the \emph{state space} and \emph{action space} respectively.
The transition probability kernel gives rise to the \emph{state transition kernel} $\cSP(s' | s,a) := \cP(\{ s' \} \times \mathbb{R} \cdbar s,a)$, which gives the probability of transitioning from state $s$ to state $s'$ if action $a$ is taken in $s$.

An agent's behavior is determined by a \emph{policy}, that defines, for each state $s\in\cS$ and time $t \in \mathbb{N}$, a probability measure over $\cA$ denoted by $\pi_t(\cdot \cdbar s)$.
A \emph{stationary policy} is a policy which is independent of time, which we will denote by $\pi(\cdot \cdbar s)$ where appropriate. 
At each time $t$, the agent communicates an action $A_{t} \sim \pi_t(\cdot \cdbar S_{t-1})$ to the system in state $S_{t-1} \in \cS$.
The system then responds with a state-reward pair $(S_{t},R_{t}) \sim \cP(\cdot \cdbar S_{t-1}, A_{t})$.
Here we will assume that each reward is bounded between $[r_{\min},r_{\max}] \subset \mathbb{R}$ and that the system starts in a state $s_0$ and executes for an infinite number of steps.
Thus the execution of the system can be described by a sequence of random variables $A_1, S_{1}, R_{1}, A_2, S_2, R_2, ..$.

The finite $m$-horizon \emph{return} from time $t$ is defined as $Z_{t} := \sum_{i=t}^{t+m-1} R_i$.
The expected $m$-horizon return from time $t$, also known as the \emph{value function}, is denoted by $V^{\pi}(s_t) := \mathbb{E}[Z_{t+1} \cdbar S_t =s_t]$.
The return space $\cZ$ is the set of all possible returns.
The \emph{action-value function} is defined by $Q^{\pi}(s_t, a_{t+1}) := \mathbb{E}[Z_{t+1} \cdbar S_t =s_t, A_{t+1}=a_{t+1}]$.
An \emph{optimal policy}, denoted by $\pi^*$, is a policy that maximizes the expected return $\mathbb{E}\left[ Z_{t+1} \cdbar S_t \right]$ for all $t$; in our setting, a state-dependent deterministic optimal policy always exists.

\subsection{Compression and Sequential Prediction }
\label{sec:compression_prediction}

We now review sequential probabilistic prediction in the context of statistical data compression. 
An alphabet $\cX$ is a set of symbols.
A string of data $x_1x_2 \ldots x_n \in \cX^n$ of length $n$ is denoted by $x_{1:n}$.
The prefix $x_{1:j}$ of $x_{1:n}$, $j\leq n$, is denoted by $x_{\leq j}$ or $x_{< j+1}$.
The empty string is denoted by $\epsilon$.
The concatenation of two strings $s$ and $r$ is denoted by $sr$.

A coding distribution $\rho$ is a sequence of probability mass functions $\rho_n : \cX^n \to [0,1]$, which for all $n\in\mathbb{N}$ satisfy the constraint that 
$\rho_n(x_{1:n}) = \sum_{y\in\cX} \rho_{n+1}(x_{1:n}y)$
for all $x_{1:n} \in \cX^n$, with the base case $\rho_0(\epsilon) := 1$.
From here onwards, whenever the meaning is clear from the argument to $\rho$, the subscript on $\rho$ will be dropped.
Under this definition, the conditional probability of a symbol $x_n$ given previous data $x_{<n}$ is defined as $\rho(x_n | x_{<n}) := \rho(x_{1:n}) / \rho(x_{<n})$ provided $\rho(x_{<n}) > 0$, with the familiar chain rules $\rho(x_{1:n}) = \prod_{i=1}^n \rho(x_i | x_{<i})$ and $\rho(x_{j:k} \cdbar x_{<j}) = \prod_{i=j}^k \rho(x_i | x_{<i})$ now following.

A binary source code $c : \cX^* \to \{ 0, 1 \}^*$ assigns to each possible data sequence $x_{1:n}$ a binary codeword $c(x_{1:n})$ of length $\ell_c(x_{1:n})$.
The typical goal when constructing a source code is to minimize the lengths of each codeword while ensuring that the original data sequence $x_{1:n}$ is always recoverable from $c(x_{1:n})$.
A fundamental technique known as \emph{arithmetic encoding} \citep{Witten87} makes explicit the connection between coding distributions and source codes.
Given a coding distribution $\rho$ and a data sequence $x_{1:n}$, arithmetic encoding constructs a code $a_{\rho}$ which produces a binary codeword whose length is essentially $-\log_2 \rho(x_{1:n})$.   
We refer the reader to the standard text of \citet{coverthomas} for further information.

\subsection{Compression-based classification}
\label{sec:compr_class}

Compression-based classification was introduced by \citet{Frank00textcategorization}.
Given a sequence of $n$ labeled i.i.d. training examples $\cD := (y_1, c_1),  \dots, (y_n, c_n)$, where $y_i$ and $c_i$ are the input and class labels respectively, one can apply Bayes rule to express the probability of a new example $Y$ being classified as class $C \in \cC$ given the training examples $\cD$ by
\begin{equation} 
\mathbb{P}\left[~C ~|~ Y, \cD~ \right] = \frac{\mathbb{P}\left[~Y ~|~ C, \cD~\right] ~ \mathbb{P}\left[~ C ~|~ \cD~\right]}
{ \sum\limits_{c \in \cC } \mathbb{P}\left[~Y ~|~ c, \cD~\right] ~ \mathbb{P}\left[~ c ~|~ \cD~\right]}.
\end{equation}
The main idea behind compression-based classification is to model $\mathbb{P}\left[~Y ~|~ C, \cD~\right]$ using a coding distribution for the inputs that is trained on the subset of examples from $\cD$ that match class $C$.
Well known non-probabilistic compression methods such as \lempelziv{} \citep{ziv78} can be used by forming their associated coding distribution $2^{-{\ell_z(x_{1:n})}}$, where $\ell_z(x_{1:n})$ is the length of the compressed data $x_{1:n}$ in bits under compression method $z$.
The class probability $\mathbb{P}\left[~ C ~|~ \cD \right]$ can be straightforwardly estimated from its empirical frequency or smoothed versions thereof.
Thus the overall accuracy of the classifier essentially depends upon how well the inputs can be modeled by the class conditional coding distribution. 

Compression-based classification has both advantages and disadvantages.
On one hand, it is straightforward to apply generic compression techniques (including those operating at the bit or character level) to complicated input types such as richly formatted text or DNA strings \citep{Frank00textcategorization,Bratko06spamfiltering}.
On the other hand, learning a probabilistic model of the input may be significantly more difficult than directly applying standard discriminative classification techniques.
Our approach to policy evaluation, which we now describe, raises similar questions.

\section{Compression and Control}

We now introduce \emph{Compress and Control} ({\sc cnc}), our new method for policy evaluation.

\subsection{Overview}

Policy evaluation is concerned with the estimation of the state-action value function $Q^{\pi}(s,a)$.
Here we assume that the environment is a finite, time homogenous MDP $\cM\nobreak:=\nobreak(\cS, \cA, \mu)$, and that the policy to be evaluated is a stationary Markov policy $\pi$.
To simplify the exposition, we consider the finite $m$-horizon case, and assume that all rewards are drawn from a finite set $\cR \subset \mathbb{R}$; later we will discuss how to remove these restrictions.

At a high level, \cnc{} performs policy evaluation by learning a \emph{time-independent} state-action conditional distribution $\prob(Z \cbar S, A)$; the main technical component of our work involves establishing that this time-independent conditional probability is well defined.
Our technique involves constructing a particular kind of \emph{augmented} Markov chain whose stationary distribution allows for the recovery of $\prob(Z \cbar S, A)$.
Given this distribution, we can obtain 
\begin{equation*}
Q^{\pi}(s,a) = \sum\limits_{z \in \cZ} z ~ \prob(Z = z \cbar S=s, A=a). 
\end{equation*}
In the spirit of compression-based classification, \cnc{} estimates this distribution by using Bayes rule to combine learnt density models of both $\prob(S \cbar Z, A)$ and $\prob(Z \cbar A)$.
Although it might seem initially strange to learn a model that conditions on the future return, the next section shows how this counterintuitive idea can be made rigorous.

\subsection{Transformation}\label{subsec:transmogrification}

Our goal is to define a transformed chain whose stationary distribution can be marginalized to obtain a distribution over states, actions and the $m$-horizon return.
We need two lemmas for this purpose.
To make these statements precise, we will use some standard terminology from the Markov chain literature; for more detail, we recommend the textbook of \citet{bremaud99}. 
\begin{defn}
A Homogenous Markov Chain (HMC) given by $\{X_t\}_{t \geq 0}$ over state space $\cX$ is said to be:
(AP) aperiodic iff \emph{\textrm{gcd}}$\{n\geq 1: \prob[X_n=x|X_0=x]>0\}=1, \forall x\in\cX$;
(PR) positive recurrent iff $\E[\min\{n\geq 1:X_n=x\}|X_0=x]<\infty, \forall x\in\cX$;
(IR) irreducible iff $\forall x,x'\,\exists n\geq 1: \prob[X_n=x'|X_0=x]>0$;
(EA) essentially aperiodic iff \emph{\textrm{gcd}}$\{n\geq 1:\prob[X_n=x|X_0=x]>0\}\in\{1,\infty\}, \forall x\in\cX$.
Note also that EA+IR implies AP.
\end{defn}
\noindent Although the term \emph{ergodic} is sometimes used to describe particular combinations of these properties (e.g. AP+PR+IR), here we avoid it in favor of being more explicit.

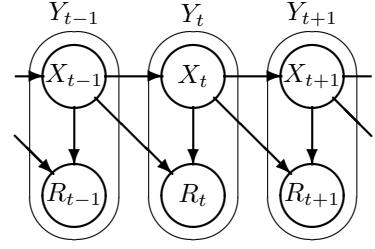
\begin{figure}[t!]
\begin{center}
\unitlength=2.5ex
\begin{picture}(12,8)
\thicklines
\put(2,5){\circle{2}}\put(2,5){\makebox(0,0)[cc]{$X_{t-1}$}}
\put(6,5){\circle{2}}\put(6,5){\makebox(0,0)[cc]{$X_t$}}
\put(10,5){\circle{2}}\put(10,5){\makebox(0,0)[cc]{$X_{t+1}$}}
\put(2,1){\circle{2}}\put(2,1){\makebox(0,0)[cc]{$R_{t-1}$}}
\put(6,1){\circle{2}}\put(6,1){\makebox(0,0)[cc]{$R_t$}}
\put(10,1){\circle{2}}\put(10,1){\makebox(0,0)[cc]{$R_{t+1}$}}
\put(0,5){\vector(1,0){1}}\put(3,5){\vector(1,0){2}}\put(7,5){\vector(1,0){2}}\put(11,5){\line(1,0){1}}
\put(2,4){\vector(0,-1){2}}\put(6,4){\vector(0,-1){2}}\put(10,4){\vector(0,-1){2}}
\put(0,3){\vector(1,-1){1.3}}\put(2.7,4.3){\vector(1,-1){2.6}}\put(6.7,4.3){\vector(1,-1){2.6}}\put(10.7,4.3){\line(1,-1){1.3}}
\thinlines
\put(2,3){\oval(3,7)}\put(2,6.7){\makebox(0,0)[cb]{$Y_{t-1}$}}
\put(6,3){\oval(3,7)}\put(6,6.7){\makebox(0,0)[cb]{$Y_{t}$}}
\put(10,3){\oval(3,7)}\put(10,6.7){\makebox(0,0)[cb]{$Y_{t+1}$}}
\end{picture}
\end{center}
\caption{\label{fig:lem1} Lemma \ref{lem:ext} applied to $\{\big((A_t,S_t), R_t\big)\}_{t\geq 1}$.}
\end{figure}

\begin{lem}\label{lem:ext}
Consider a stochastic process $\{ X_t \}_{t\geq1}$ over state space $\cX$ that is independent of a sequence of $\cU$-valued random variables $\{U_t\}_{t\geq1}$ in the sense that
$\prob(x_t | x_{<t}, u_{<t})=\prob(x_t|x_{<t})$, and with $U_t$ only depending on $X_{t-1}$ and $X_t$ in the sense
that $\prob(u_t|x_{1:t}, u_{<t})=\prob(u_t|x_{t-1},x_t)$
and $\prob(U_t=u|X_{t-1}=x,X_t=x')$ being independent of $t$.
Then, if $\{X_t\}_{t\geq 1}$ is an (IR/EA/PR) HMC over $\cX$,
then $\{Y_t\}_{t\geq1}:= \{ (X_t,U_t) \}_{t\geq1}$ is an (IR/EA/PR) HMC
over $\cY:=\{y_t \in \cX \times \cU : \exists x_{t-1}\in\cX : \prob(y_t|x_{t-1})>0\}$.
\end{lem}
\noindent Lemma \ref{lem:ext} allows HMC $\{X_t := (A_t, S_t) \}_{t \geq 1}$ to be augmented to obtain the HMC $\{Y_t:=(X_t, R_t)\}_{t\geq 1}$, where $A_t$, $S_t$ and $R_t$ denote the action, state and reward at time $t$ respectively; see Figure \ref{fig:lem1} for a graphical depiction of the dependence structure.

The second result allows the HMC $\{Y_t\}_{t\geq 1}$ to be further augmented to give the snake HMC $\{Y_{t:t+m}\}_{t\geq 1}$ \citep{bremaud99}.
This construction ensures that there is sufficient information within each augmented state to be able to condition on the $m$-horizon return.

\begin{lem}\label{lem:snake}
If $\{Y_t\}_{t \geq 1}$ is an (IR/EA/PR) HMC over state space $\cY$, then for any $m\in\SetN$, the stochastic process $\{ W_t \}_{t \geq 1}$, where $W_t:=(Y_t,...,Y_{t+m})$, is an (IR/EA/PR) HMC over $\cW :=\{y_{0:m}\in\cY^{m+1} : \mathbb{P}(y_{1:m}|y_0)>0\}$.
\end{lem}

Now if we assume that the HMC defined by $\cM$ and $\pi$ is (IR+EA+PR), Lemmas \ref{lem:ext} and \ref{lem:snake} imply that there exists a unique stationary distribution $\nu'$ over the augmented state space $(\cA \times \cS \times \cR)^{m+1}$.

Furthermore, if we let $( A'_0,S'_0,R'_0, \dots, A'_m,S'_m,R'_m ) \sim \nu'$ and define $Z' := \sum_{i=1}^m R'_i$, it is clear that there exists a joint distribution $\nu$ over $\cZ \times (\cA \times \cS \times \cR)^{m+1}$ such that $(Z', A'_0,S'_0,R'_0, \dots, A'_m,S'_m,R'_m) \sim \nu$.
Hence the $\nu$-probability $\prob\left[ Z' ~|~ S'_0, A'_1 \right]$ is well defined, which allows us to express the action-value function $Q^\pi$ as
\begin{equation}\label{eq:dual}
Q^\pi(s,a) = \mathbb{E}_\nu\left[ Z' ~|~ S'_0=s, A'_1=a \right].
\end{equation}

Finally, by expanding the expectation and applying Bayes rule, Equation \ref{eq:dual} can be further re-written as
\begin{eqnarray}\label{eq:bayes_sloppy}
Q^{\pi}(s,a) &=& \sum_{z \in \cZ} z ~ \nu(z ~|~ s, a) \notag \\
&=& \sum_{z \in \cZ} z ~ \frac{\nu(s ~|~ z, a) ~\nu(z ~|~ a)}{\sum\limits_{z' \in \cZ} \nu(s ~|~ z', a) ~\nu(z' ~|~ a)}.
\end{eqnarray}

The {\sc cnc} approach to policy evaluation involves directly learning the conditional distributions $\nu(s ~|~ z, a)$ and $\nu(z ~|~ a)$ in Equation \ref{eq:bayes_sloppy} from data, and then using these learnt distributions to form a plug-in estimate of $Q^\pi(s,a)$.
Notice that $\nu(s | z, a)$ conditions on the return, similar in spirit to prior work on planning as inference \citep{Attias03planningby,botvinick12,botvinick12b}. 
The distinguishing property of {\sc cnc} is that the conditioning is performed with respect to a stationary distribution that has been explicitly constructed to allow for efficient modeling and inference.

\subsection{Online Policy Evaluation}

We now provide an online algorithm for compression-based policy evaluation.
This will produce, for all times $t \in \mathbb{N}$, an estimate $\hat{Q}^{\pi}_t(s, a)$ of the $m$-horizon expected return $Q^\pi(s,a)$ as a function of the first $t-m$ action-observation-reward triples.

Constructing our estimate involves modeling the $\nu$-probability terms in Equation \ref{eq:bayes_sloppy} using two different coding distributions, $\rhos$ and $\rhoz$ respectively; $\rhos$ will encode states conditional on return-action pairs, and $\rhoz$ will encode returns conditional on actions.
Sample states, actions and returns can be generated by directly executing the system $(\cM,\pi)$; Provided the HMC $\cM+\pi$ is (IR+EA+PR), Lemmas \ref{lem:ext} and \ref{lem:snake} ensure that the empirical distributions formed from a sufficiently large sample of action/state/return triples will be arbitrarily close to the required conditional $\nu$-probabilities.

Next we describe how the coding distributions are trained. 
Given a history $h_t := s_0, a_1, s_1, r_1 \dots, a_{n+m}, s_{n+m}, r_{n+m}$ with $t=n+m$, we define the $m$-lagged return at any time $i \leq n+1$ by $z_i := r_i + \dots + r_{i+m-1}$.
The sequence of the first $n$ states occurring in $h_{t}$ can be mapped to a subsequence denoted by $s^{z,a}_{0:n-1}$ that is defined by keeping only the states  $( s_i : z_{i+1} = z \wedge a_{i+1} = a )^{n-1}_{i=0}$.
Similarly, a sequence of $m$-lagged returns $z_{1:n}$ can be mapped to a subsequence $z^{a}_{1:n}$ formed by keeping only the returns $( z_i : a_{i} = a )^n_{i=1}$ from $z_{1:n}$.
Our value estimate at time $t$ of taking action $a$ in state $s$ can now be defined as
\begin{equation}\label{eq:cnc_value_estimate}
\hat{Q}^{\pi}_t(s, a) := \sum_{z \in \cZ} z \, w^{z,a}_t(s), 
\end{equation}
where
\begin{equation}\label{eq:cnc_bayes}
w_t^{z,a}(s) := \frac{\rhos(~s \cdbar s_{0:n-1}^{z, a} ~) \, \rhoz(z \cdbar z^{a}_{1:n}) }{\sum\limits_{z' \in \cZ} \rhos(s \cdbar s_{0:n-1}^{z', a}) \, \rhoz(z' \cdbar z^{a}_{1:n})  }
\end{equation}
approximates the probability of receiving a return of $z$ if action $a$ is selected in state $s$.

\paragraph{Implementation.}

\begin{algorithm}[t!]
\caption{{\sc cnc policy evaluation}\label{alg:cnc}}
\begin{algorithmic}[1]
\medskip
\REQUIRE Stationary policy $\pi$, environment $\cM$ 
\REQUIRE Finite planning horizon $m \in \mathbb{N}$
\REQUIRE Coding distributions $\rhos$ and $\rhoz$
\medskip
\FOR{$i=1$ to $t$}
	  \STATE Perform $a_i \sim \pi(\cdot ~|~ s_{i-1})$
		\STATE Observe $(s_i, r_i) \sim \mu(\cdot ~|~ s_{i-1}, a_i)$
		\IF{$i \geq m$}
		    \STATE \text{Update} $\rhos$ \text{in bucket} $(z_{i-m+1}, a_{i-m+1})$ \text{with} $s_{i-m}$
				\STATE \text{Update} $\rhoz$ \text{in bucket} $a_{i-m+1}$ \text{with} $z_{i-m+1}$
		\ENDIF
\ENDFOR
\medskip
\STATE \bf{return} $\hat{Q}_{t}^\pi$
\end{algorithmic}
\end{algorithm}

The action-value function estimate $\hat{Q}_t^\pi$ can be computed efficiently by maintaining $|\cZ||\cA|$ \emph{buckets}, each corresponding to a particular return-action pair $(z,a)$. Each bucket contains an instance of the coding distribution $\rhos$ encoding the state sequence $s^{z,a}_{0:n-1}$. 
Similarly, $|\cA|$ buckets containing instances of $\rhoz$ are created to encode the various return subsequences.
This procedure is summarized in Algorithm \ref{alg:cnc}.

To obtain a particular state-action value estimate, Equations \ref{eq:cnc_value_estimate} and \ref{eq:cnc_bayes} can be computed directly by querying the appropriate bucketed coding distributions.
Assuming that the time required to compute each conditional probability using $\rhos$ and $\rhoz$ is constant, the time complexity for computing $\hat{Q}_t(s,a)$  is $O(|\cZ|)$.

\subsection{Analysis}
\label{sec:analysis}

We now show that the state-action estimates defined by Equation \ref{eq:cnc_value_estimate} are consistent provided that consistent density estimators are used for both $\rhos$ and $\rhoz$.
Also, we will say $f_n$ converges stochastically to 0 with rate $n^{-1/2}$ if and only if
\begin{equation*}
\label{Opdef}
  \exists c\!>\!0,~\forall\delta\!\in\![0,1] : \prob \Big(|f_n(\omega)|\leq\sqrt{\tfrac{c}{n}\ln\fr2\delta}~\Big) \geq 1-\delta,
\end{equation*}
and will denote this by writing $f_n(\omega) \in O_\prob(n^{-1/2})$. 
\ifextended
Further detail regarding the results in this section can be found in the appendix.
\fi
\begin{thm}\label{thm:convergence}
Given an $m$-horizon, finite state space, finite action space, time homogenous MDP $\cM := (\cS, \cA, \mu)$ and a stationary policy $\pi$ that gives rise to an (IR+EA+PR) HMC, for all $\epsilon > 0$, we have that for any state $s\in\cS$ and action $a \in \cA$ that
\begin{equation*}
\lim_{n \to \infty} \mathbb{P} \left[ ~|~ \hat{Q}^{\pi}_n(s, a) - Q^{\pi}(s,a) ~|~ \geq \epsilon \right] = 0,
\end{equation*}
provided $\rhos$ and $\rhoz$ are consistent estimators of $\nu(s | z, a)$ and $\nu( z | a)$ respectively.
Furthermore, if $|\rhos(s | z,a) - \nu(s | z, a)| \in O_\prob(n^{-1/2})$ and 
$|\rhoz(z | a) - \nu(z | a)| \in O_\prob(n^{-1/2})$
then $|\hat{Q}^\pi_n(s,a) - Q^\pi(s,a)| \in O_\prob(n^{-1/2})$.
\end{thm}
\ifextended
\begin{proof}
See the appendix.
\end{proof}
\fi

Next we state consistency results for two types of estimators often used in model-based reinforcement learning. 

\begin{thm}
The frequency estimator
$\rho(x_n | x_{<n}) := \tfrac{1}{n-1} \sum_{i=1}^{n-1} \llbracket x_n = x_i \rrbracket$
when used as either $\rhos$ or $\rhoz$ is a consistent estimator of $\nu(s|z,a)$ or $\nu(z|a)$ respectively for any $s \in \cS$, $z \in \cZ$, and $a \in \cA$; furthermore, the absolute estimation error converges stochastically to 0 with rate $n^{-1/2}$.
\end{thm}
Note that the above result is essentially tabular, in the sense that each state is treated atomically.
The next result applies to a factored application of multi-alphabet Context Tree Weighting ({\sc ctw}) \citep{tjalkens93,ctw95,veness10b}, which can handle considerably larger state spaces in practice.
In the following, we use the notation $s_{n,i}$ to refer to the $i$th factor of state $s_n$.

\begin{thm}
Given a state space that is factored in the sense that $\cS~:=~\cB_1 \times \dots \times \cB_k$, the estimator $\rho(s_n \,|\, s_{<n}) := \prod_{i=1}^k \text{\sc ctw}(s_{n,i} ~|~ s_{n,<i}, s_{<n,1:i})$ when used as $\rhos$, is a consistent estimator of $\nu(s | z, a)$ for any $s \in \cS$, $z \in \cZ$, and $a \in \cA$; furthermore, the absolute estimation error converges stochastically to 0 at a rate of $n^{-1/2}$.
\end{thm}

\section{Experimental Results}
\label{sec:experiments}

In this section we describe two sets of experiments.
The first set is an experimental validation of our theoretical results using a standard policy evaluation benchmark. 
The second combines {\sc cnc} with a variety of density estimators and studies the resulting behavior in a large on-policy control task. 

\subsection{Policy Evaluation}

Our first experiment involves a simplified version of the game of Blackjack \citep[Section 5.1]{Sutton:1998}.
In Blackjack, the agent requests cards from the dealer.
A game is won when the agent's card total exceeds the dealer's own total. 
We used {\sc cnc} to estimate the value of the policy that stays if the player's sum is 20 or 21, and hits in all other cases. 
A state is represented by the single card held by the dealer, the player's card total so far, and whether the player holds a usable ace. 
In total, there are 200 states, two possible actions (hit or stay), and three possible returns (-1, 0 and 1).
A Dirichlet-Multinomial model with hyper-parameters $\alpha_i=\tfrac{1}{2}$ was used for both $\rhos$ and $\rhoz$. 

\begin{figure}[t!]
\centerline{
\includegraphics[width=1.65in]{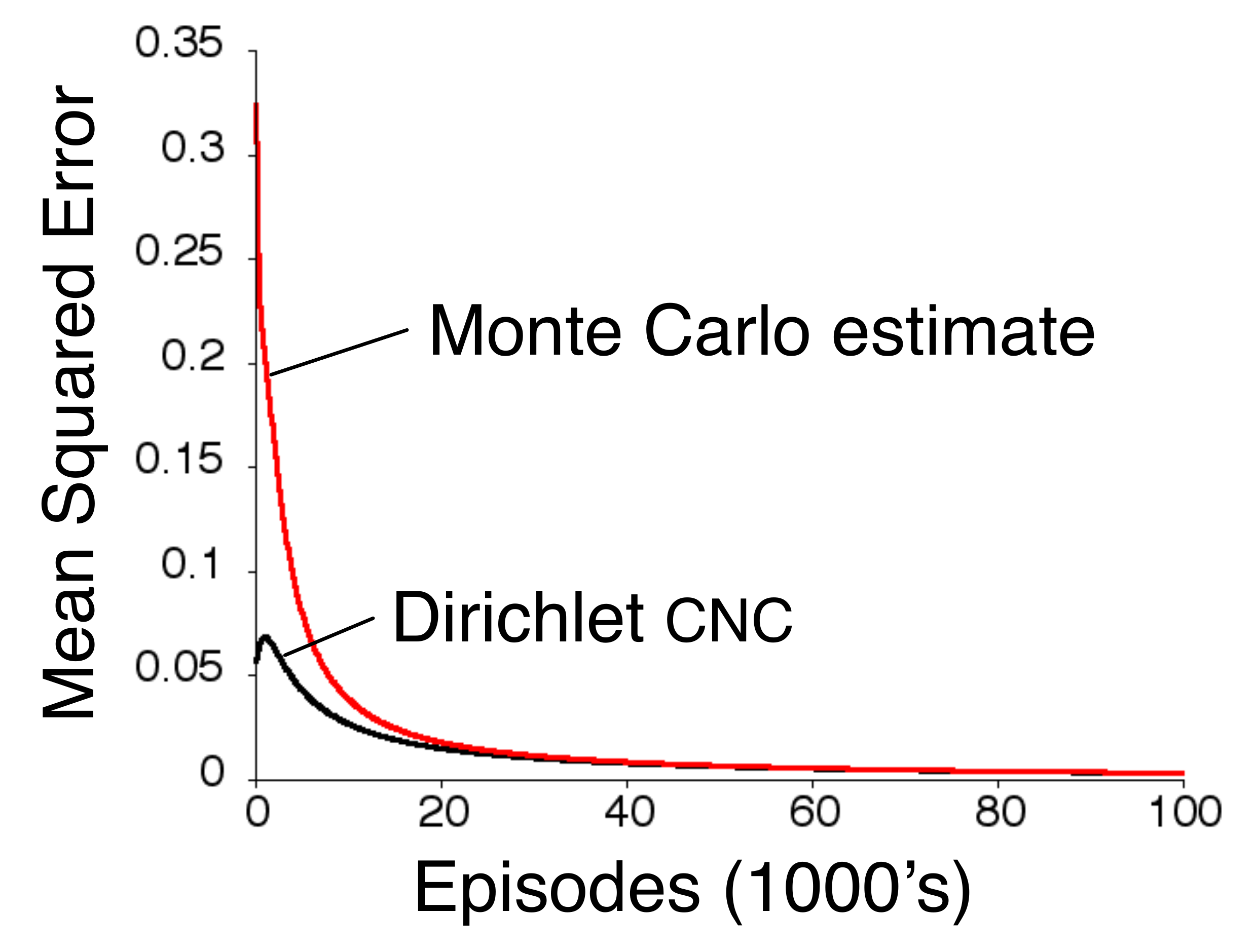}
\includegraphics[width=1.65in]{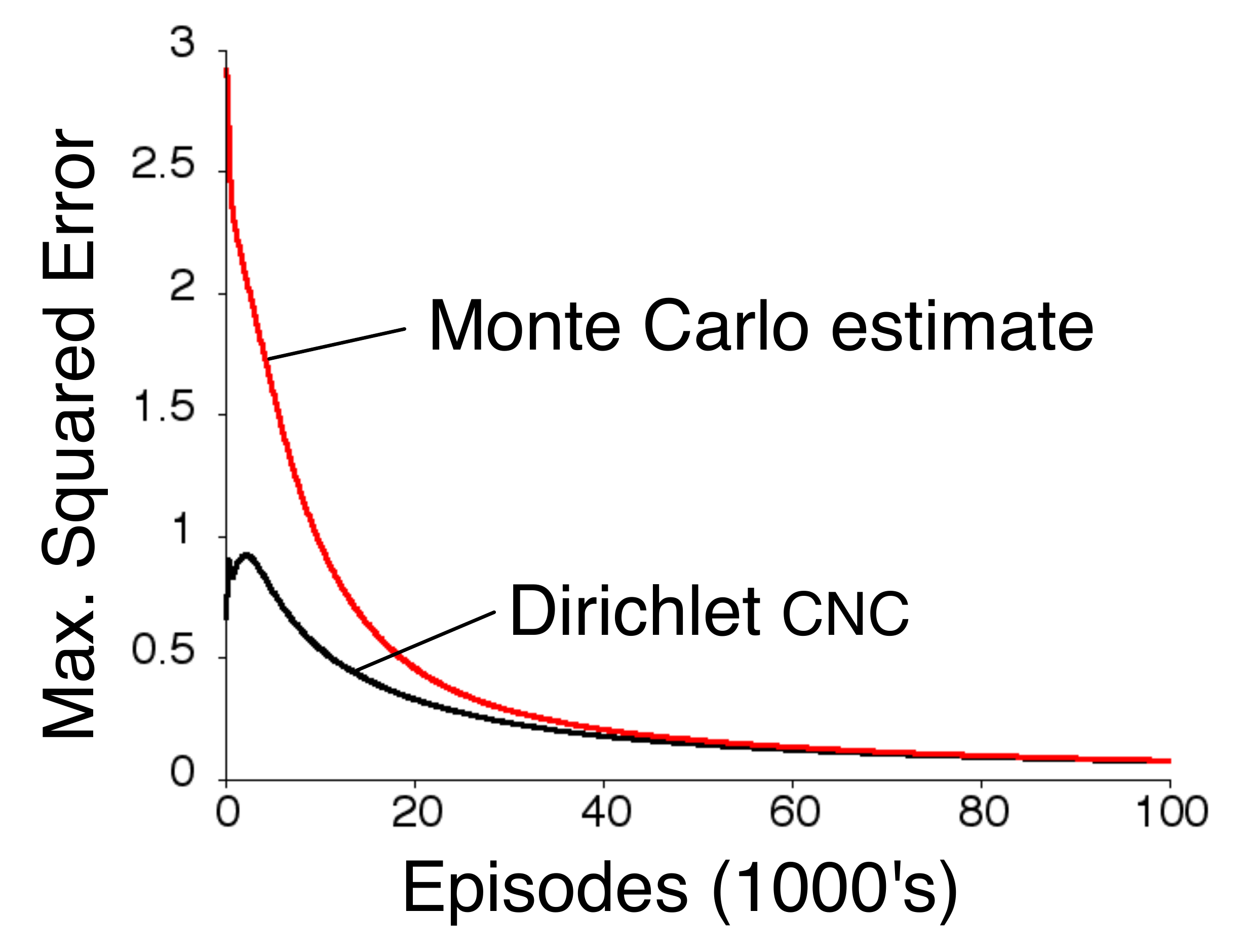}
}
\caption{Mean and maximum squared errors of the Monte Carlo and \cnc{} estimates on the game of Blackjack.\label{fig:blackjack-mse}}
\end{figure}

Figure \ref{fig:blackjack-mse} depicts the estimated MSE and average maximum squared error of $\hat{Q}^\pi$ over 100,000 episodes; the mean and maximum are taken over all possible state-action pairs and averaged over 10,000 trials. 
We also compared {\sc cnc} to a first-visit Monte Carlo value estimate \citep{szepesvari10}. 
The {\sc cnc} estimate closely tracks the Monte Carlo estimate, even performing slightly better early on due to the smoothing introduced by the use of a Dirichlet prior.
As predicted by the analysis in Section \ref{sec:analysis}, the MSE decays toward zero.

\subsection{On-policy Control} 

Our next set of experiments explored the on-policy control behavior of {\sc cnc} under an $\epsilon$-greedy policy.
The purpose of these experiments is to demonstrate the potential of {\sc cnc} to scale to large control tasks when combined with a variety of different density estimators.
Note that Theorem \ref{thm:convergence} does not apply here: using \cnc{} in this way violates the assumption that $\pi$ is stationary. 

\subsubsection{Evaluation Platform.}

We evaluated \cnc{} using ALE, the Arcade Learning Environment \citep{BellemareNVB13}, a reinforcement learning interface to the Atari 2600 video game platform. 
Observations in ALE consist of frames of $160 \times 210$ 7-bit color pixels generated by the Stella Atari 2600 emulator. 
Although the emulator generates frames at 60Hz, in our experiments we consider time steps that last 4 consecutive frames, following the existing literature \citep{bellemare14skip,mnih13playing}.
We first focused on the game of \textsc{Pong}, which has an action space of $\{ \textsc{Up}, \textsc{Down}, \textsc{Noop} \}$ and provides a reward of 1 or -1 whenever a point is scored by either the agent or its computer opponent. 
Episodes end when either player has scored 21 points; as a result, possible scores for one episode range between -21 to 21, with a positive score corresponding to a win for the agent. 

\subsubsection{Experimental Setup.}

We studied four different {\sc cnc} agents, with each agent corresponding to a different choice of model for $\rhos$; the Sparse Adapative Dirichlet ({\sc sad}) estimator \citep{hutter13sparse} was used for $\rhoz$ for all agents.
Each agent used an $\epsilon$-greedy policy \citep{Sutton:1998} with respect to its current value function estimates.
The exploration rate $\epsilon$ was initialized to 1.0, then decayed linearly to 0.02 over the course of 200,000 time steps. 
The horizon was set to $m=80$ steps, corresponding to roughly 5 seconds of play. 
The agents were evaluated over 10 trials, each lasting 2 million steps.

The first model we consider is a factored application of the \sad{} estimator, a count based model designed for large, sparse alphabets.
The model divides the screen into $16\times16$ regions.  
The probability of a particular image patch occurring within each region is modeled using a region-specific \sad{} estimator.
The probability assigned to a whole screen is the product of the probabilities assigned to each patch.

The second model is an auto-regressive application of logistic regression \citep{bishop2006}, that assigns a probability to each pixel using a shared set of parameters.
The product of these per-pixel probabilities determines the probability of a screen under this model.
The features for each pixel prediction correspond to the pixel's local context, similar to standard context-based image compression techniques \citep{witten99managing}.
The model's parameters were updated online using {\sc Adagrad} \citep{Duchi:2011}.
The hyperparameters (including learning rate, choice of context, etc.) were optimized via the random sampling technique of \citet{Bergstra:2012}.

The third model uses the \lempelziv{} algorithm \citep{ziv78}, a dictionary-based compression technique.
It works by adapting its internal data structures over time to assign shorter code lengths to more frequently seen substrings of data.
For our application, the pixels in each frame were encoded in row-major order, by first searching for the longest sequence in the history matching the new data to be compressed, and then encoding a triple that describes the temporal location of the longest match, its length, as well as the next unmatched symbol.
This process repeats until no data is left.
Recalling Section \ref{sec:compr_class}, the (implicit) conditional probability of a state $s$ under the \lempelziv{} model can now be obtained by computing
\begin{equation*}
\rhos(s \,|\, s^{z,a}_{0:n-1}) := 2^{- \left[ \ell_{\sc LZ}(s^{z,a}_{0:n-1} s) - \ell_{\sc LZ}(s^{z,a}_{0:n-1}) \right]} .
 \end{equation*}

\begin{figure}
\centerline{
\includegraphics[width=1.65in]{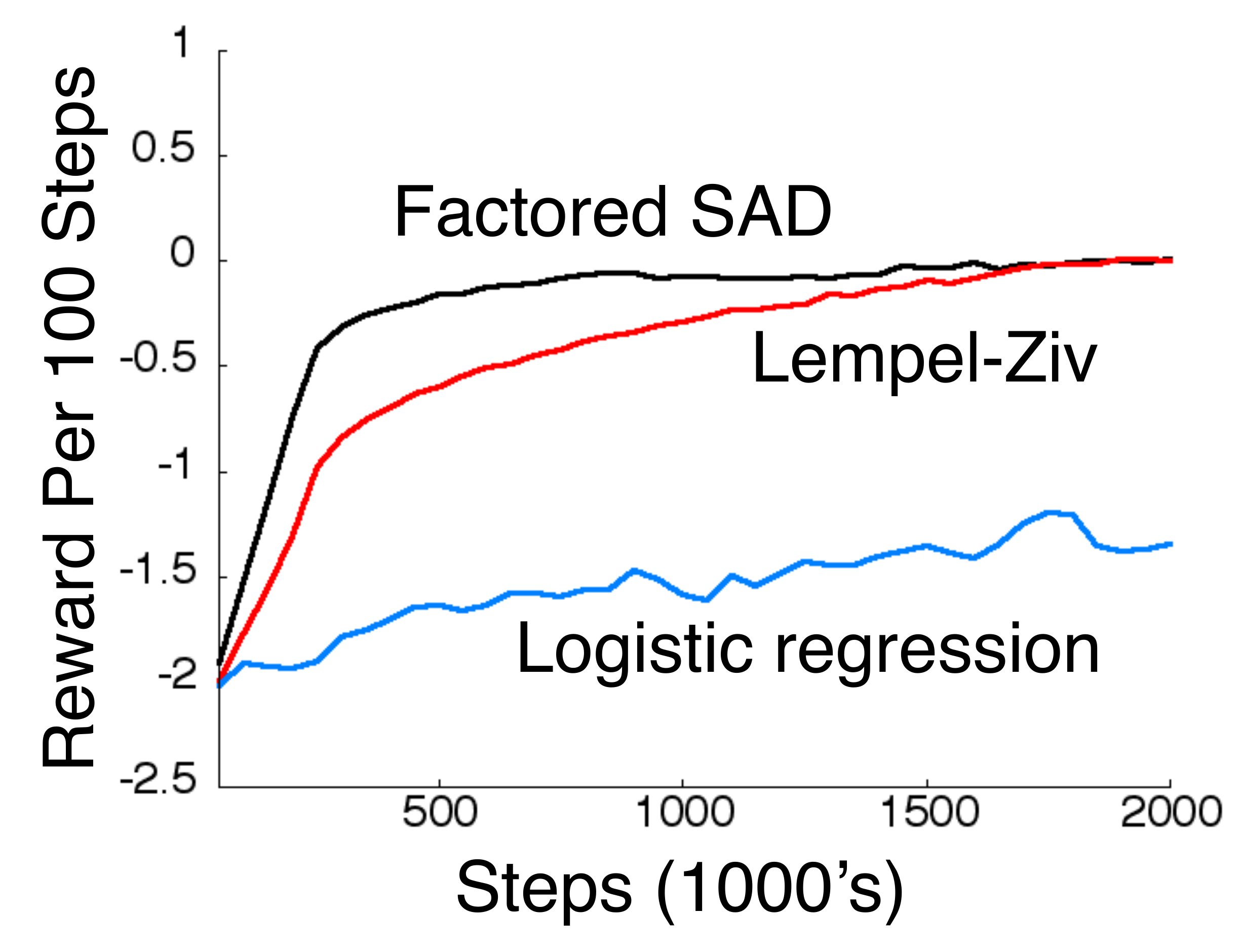}
\includegraphics[width=1.65in]{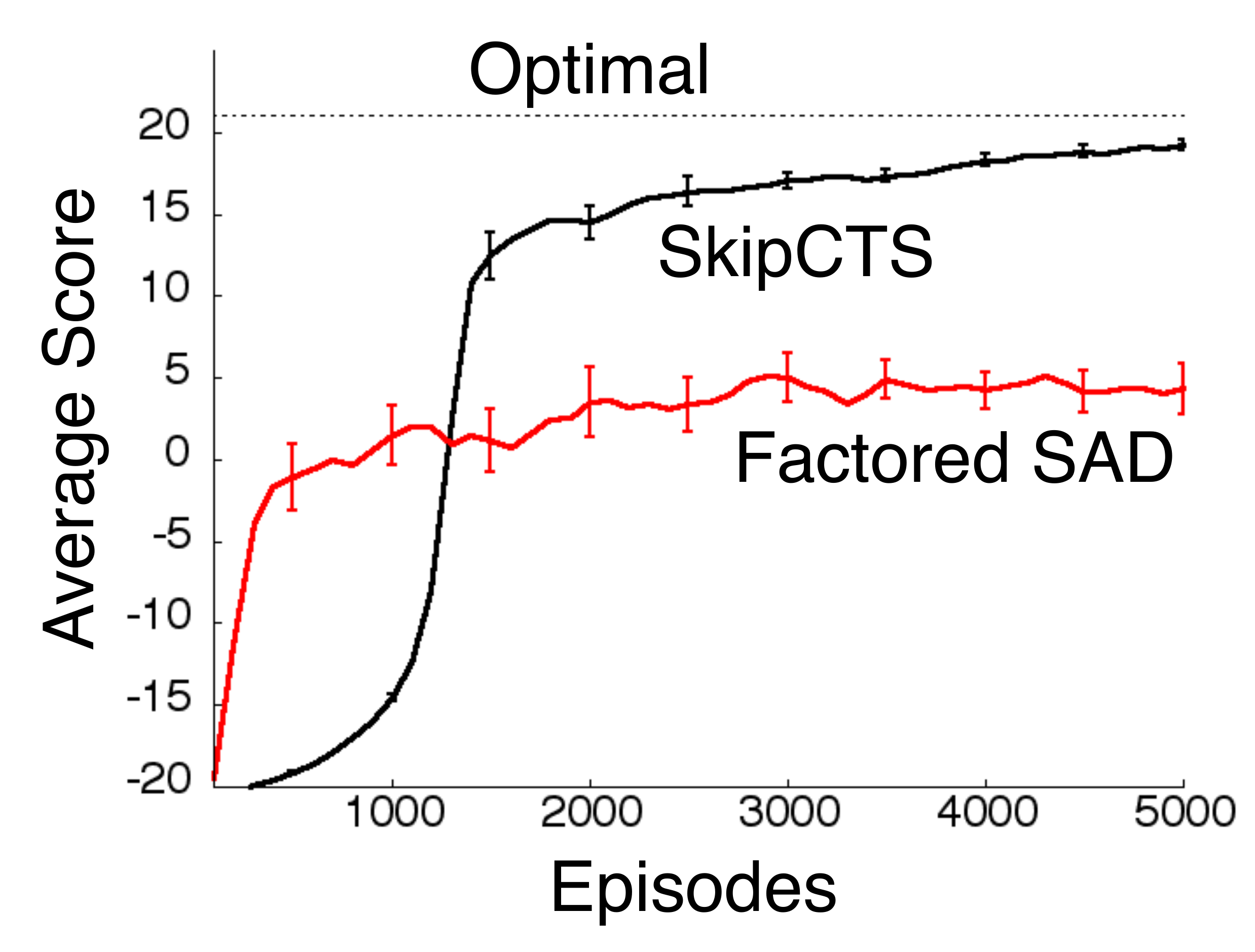}
}
\caption{\emph{Left.} Average reward over time in \pong{}. \emph{Right.} Average score across episodes in \pong{}. Error bars indicate one inter-trial standard error.\label{fig:diptych}}
\end{figure}

\subsubsection{Results.} 

As depicted in Figure \ref{fig:diptych} (left), all three models improved their policies over time. 
By the end of training, two of these models had learnt control policies achieving win rates of approximately 50\% in \pong{}. 
Over their last 50 episodes of training, the \lempelziv{} agents averaged -0.09 points per episode (std. error: 1.79) and the factored SAD agents, 3.29 (std. error: 2.49).
While the logistic regression agents were less successful (average -17.87, std. error 0.38) we suspect that further training time would significantly improve their performance.
Furthermore, all agents ran at real-time or better.
These results highlight how \cnc{} can be successfully combined with fundamentally different approaches to density estimation.

\begin{figure}
\centerline{
\includegraphics[width=2.3in]{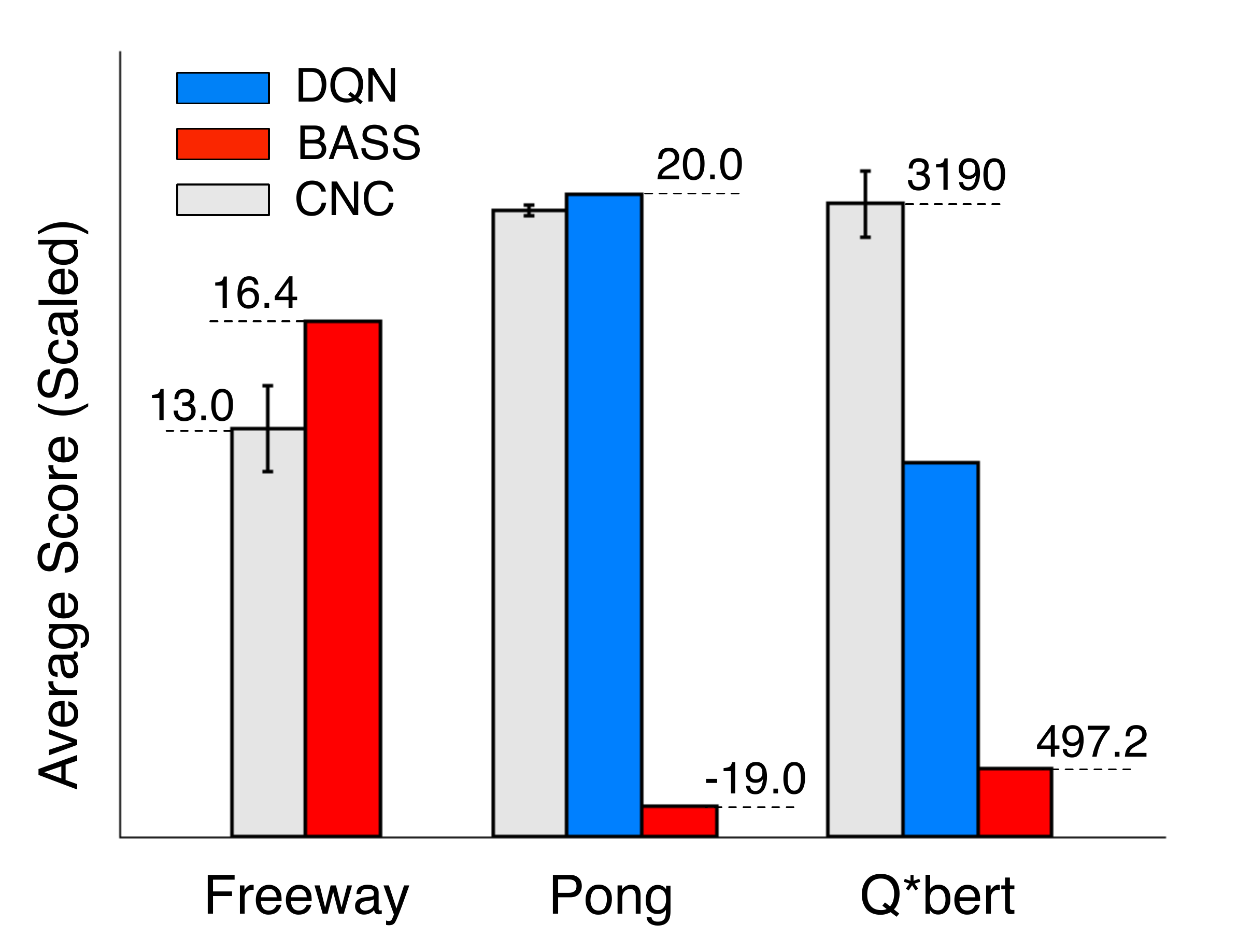}
}
\caption{Average score over the last 500 episodes for three Atari 2600 games. Error bars indicate one inter-trial one standard error.\label{fig:cool_histogram}} 
\end{figure}

We performed one more experiment to illustrate the effects of combining \cnc{} with a more sophisticated density model.
We used \skipcts{}, a recent Context Tree Weighting derivative, with a context function tailored to the ALE observation space \citep{bellemare14skip}. 
As shown in Figure \ref{fig:diptych} (right), \cnc{} combined with \skipcts{} learns a near-optimal policy in \pong{}.
We also compared our method to existing results from the literature \citep{BellemareNVB13,mnih13playing}, although note that the {\sc DQN} scores, which correspond to a different training regime and do not include Freeway, are included only for illustrative purposes.
As shown in Figure \ref{fig:cool_histogram}, \cnc{} can also learn competitive control policies on {\sc Freeway} and {\sc Q*bert}.

Interestingly, we found \skipcts{} to be insufficiently accurate for effective MCTS planning when used as a forward model, even with enhancements such as double progressive widening \citep{coutoux11}.
In particular, our best simulation-based agent did not achieve a score above $-14$ in \pong{}, and performed no better than random in {\sc Q*bert} and {\sc Freeway}.
In comparison, our {\sc cnc} variants performed significantly better using orders of magnitude less computation.
While it would be premature to draw any general conclusions, the {\sc cnc} approach does appear to be more forgiving of modeling inaccuracies. 

\section{Discussion and Limitations}

The main strength and key limitation of the {\sc cnc} approach seems to be its reliance on an appropriate choice of density estimator.
One could only expect the method to perform well if the learnt models can capture the observational structure specific to high and low return states.
Specifying a model can be thus viewed as committing to a particular kind of compression-based similarity metric over the state space.
The attractive part of this approach is that density modeling is a well studied area, which opens up the possibility of bringing in many ideas from machine learning, statistics and information theory to address fundamental questions in reinforcement learning.
The downside of course is that density modeling is itself a difficult problem.
Further investigation is required to better understand the circumstances under which one would prefer {\sc cnc} over more traditional model-free approaches that rely on function approximation to scale to large and complex problems.

So far we have only applied \cnc\ to undiscounted, finite horizon problems with finite action spaces, and more importantly, finite (and rather small) return spaces.
This setting is favorable for \cnc, since the per-step running time depends on $|\cZ| \leq m|\rmax - \rmin|$; in other words, the worst case running time scales no worse than linearly in the length of the horizon.
However, even modest changes to the above setting can change the situation drastically. 
For example, using discounted return can introduce an exponential dependence on the horizon.
Thus an important topic for future work is to further develop the {\sc cnc} approach for large or continuous return spaces.
Since the return space is only one dimensional, it would be natural to consider various discretizations of the return space.
For example, one could consider a tree based discretization that recursively subdivides the return space into successively smaller halves.
A binary tree of depth $d$ would produce $2^d$ intervals of even size with an accuracy of $\epsilon = m (\rmax - \rmin) / 2^d$. 
This implies that to achieve an accuracy of at least $\epsilon$ we would need to set  
$d \geq \log_2 \left( m (\rmax - \rmin) / \epsilon \right)$, which should be feasible for many applications.
Furthermore, one could attempt to adaptively learn the best discretization \citep{Hutter:05bayestree} or approximate Equation \ref{eq:cnc_value_estimate} using Monte Carlo sampling. 
These enhancements seem necessary before we could consider applying {\sc cnc} to the complete suite of ALE games.

\section{Closing Remarks}

This paper has introduced {\sc cnc}, an information-theoretic policy evaluation and on-policy control technique for reinforcement learning.
The most interesting aspect of this approach is the way in which it uses a learnt probabilistic model that conditions on the future return; remarkably, this counterintuitive idea can be justified both in theory and in practice.

While our initial results show promise, a number of open questions clearly remain.
For example, so far the {\sc cnc} value estimates were constructed by using only the Monte Carlo return as the learning signal.
However, one of the central themes in Reinforcement Learning is \emph{bootstrapping}, the idea of constructing value estimates on the basis of other value estimates \citep{Sutton:1998}.
A natural question to explore is whether bootstrapping can be incorporated into the learning signal used by {\sc cnc}.

For the case of on-policy control, it would be also interesting to investigate the use of compression techniques or density estimators that can automatically adapt to non-stationary data.
A promising line of investigation might be to consider the class of meta-algorithms given by \citet{gyorgy12}, that can convert any stationary coding distribution into its piece-wise stationary extension; efficient algorithms from this class have shown promise for data compression applications, and come with strong theoretical guarantees \citep{ptw}.
Furthermore, extending the analysis in Section \ref{sec:analysis} to cover the case of on-policy control or policy iteration \citep{howard1960dynamic} would be highly desirable.

Finally, we remark that information-theoretic perspectives on reinforcement learning have existed for some time; in particular, \citet{Hutter:04uaibook} described a unification of algorithmic information theory and reinforcement learning, leading to the AIXI optimality notion for reinforcement learning agents.
Establishing whether any formal connection exists between this body of work and ours is deferred to the future.

\subsubsection*{Acknowledgments.}

We thank Kee Siong Ng, Andras Gy\"{o}rgy, Shane Legg, Laurent Orseau and the anonymous reviewers for their helpful feedback on earlier revisions.

\bibliographystyle{aaai}
\bibliography{compression_to_control}

\end{document}